\documentclass[10pt,twocolumn,letterpaper]{article}

\usepackage[pagenumbers]{cvpr} 

\usepackage{graphicx}
\usepackage{amsmath}
\usepackage{amssymb}
\usepackage{booktabs}
\usepackage{color}
\usepackage{multicol}
\usepackage{caption}
\usepackage{multirow}
\usepackage{mathtools}
\usepackage{multicol}
\usepackage{caption}
\usepackage{mathtools}

\usepackage{bm}
\usepackage[accsupp]{axessibility}

\usepackage[dvipsnames]{xcolor}
\definecolor{citec}{RGB}{21, 101, 192}
\definecolor{refc}{RGB}{220, 40, 40}
\usepackage[pagebackref=true,breaklinks=true,letterpaper=true,colorlinks,bookmarks=false,citecolor=citec,linkcolor=refc]{hyperref}

\usepackage[capitalize]{cleveref}
\crefname{section}{Sec.}{Secs.}
\Crefname{section}{Section}{Sections}
\Crefname{table}{Table}{Tables}
\crefname{table}{Tab.}{Tabs.}
\newcommand{\printfnsymbol}[1]{%
        \textsuperscript{\@fnsymbol{#1}}%
}

\begin{document}

\title{
DreamSpace: Dreaming Your Room Space with Text-Driven \\ Panoramic Texture Propagation
}

\author{
Bangbang Yang$^{1}$ \quad
Wenqi Dong$^{2}$  \quad
Lin Ma$^{1}$ \quad
Wenbo Hu$^{1}$ \\
Xiao Liu$^{1}$ \quad
Zhaopeng Cui$^{2}$ \quad
Yuewen Ma$^{1}$\footnotemark[1] \\
$^{1}$PICO, ByteDance \quad
$^{2}$State Key Lab of CAD\&CG, Zhejiang University
}

\twocolumn[{%
\renewcommand\twocolumn[1][]{#1}%
\maketitle
\begin{center}
    \centering
    \includegraphics[width=1.0\linewidth, trim={0 0 0 0}, clip]{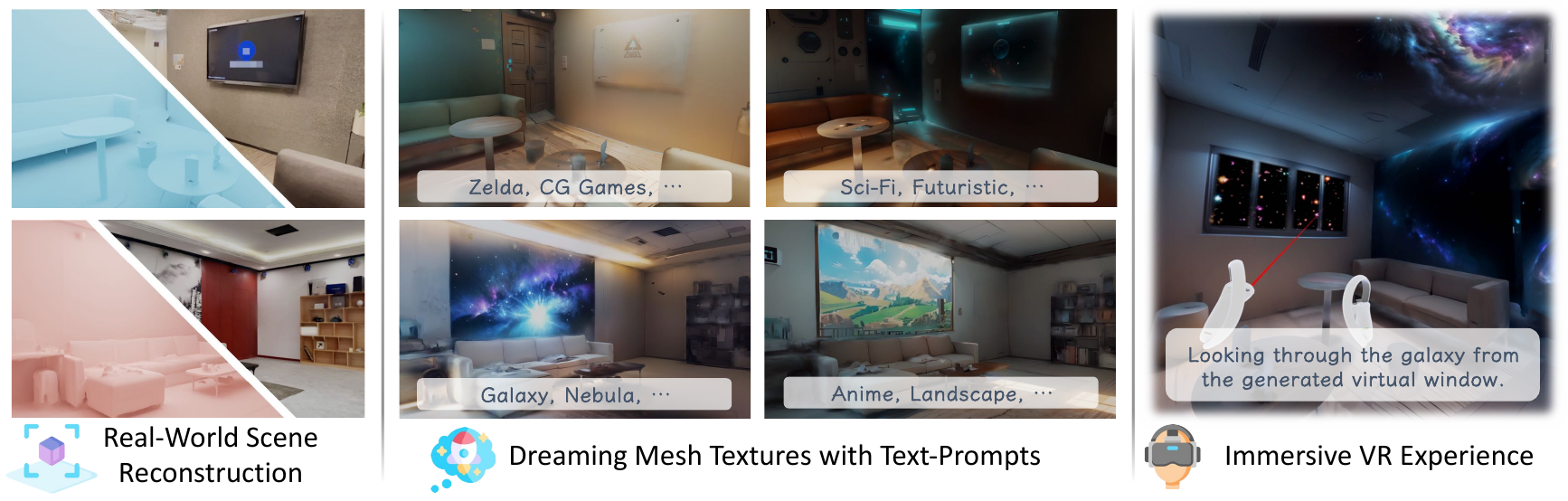}
    \captionof{figure}{
DreamSpace allows users to personalize their own spaces' appearances with text prompts and delivers immersive VR experiences on HMD devices.
  Specifically, given a real-world captured room, we generate enchanting and holistic mesh textures based on the user's textual inputs, while ensuring semantic consistency and spatial coherence (\eg, the sofa still retain its recognizable form as a sofa, but in fantasy styles).
    }
    \label{fig:teaser}
\end{center}%
}]
\renewcommand{\thefootnote}{\fnsymbol{footnote}}
\footnotetext[1]{Corresponding author.}

\begin{abstract}
Diffusion-based methods have achieved prominent success in generating 2D media.
However, accomplishing similar proficiencies for scene-level mesh texturing in 3D spatial applications, \eg, XR/VR, remains constrained, primarily due to the intricate nature of 3D geometry and the necessity for immersive free-viewpoint rendering.
In this paper, we propose a novel indoor scene texturing framework, which delivers text-driven texture generation with enchanting details and authentic spatial coherence.
The key insight is to first imagine a stylized 360$^{\circ}$ panoramic texture from the central viewpoint of the scene, and then propagate it to the rest areas with inpainting and imitating techniques.
To ensure meaningful and aligned textures to the scene, we develop a novel coarse-to-fine panoramic texture generation approach with dual texture alignment, which both considers the geometry and texture cues of the captured scenes.
To survive from cluttered geometries during texture propagation, we design a separated strategy, which conducts texture inpainting in confidential regions and then learns an implicit imitating network to synthesize textures in occluded and tiny structural areas.
Extensive experiments and the immersive VR application on real-world indoor scenes demonstrate the high quality of the generated textures and the engaging experience on VR headsets.
Project webpage: {\url{https://ybbbbt.com/publication/dreamspace}}.
\vspace{-1.0em}
\end{abstract}

\section{Introduction}

\maketitle

In our childhood, we might have imagined the world we live in with fantasy looking that follows real-world shapes but beyond reality, such as starry skies on the rooftops, beds with fancy adventurous decorations, or even virtual windows through which to gaze upon the galaxy.
Nowadays, with the advancements of HMD devices, we have the ability to visually immerse ourselves in virtual scenes with 6-DoF rendering, which opens up the possibility of experiencing scene assets with various stylized textures.
Consequently, a following question is: can we realize the dream of generating fully-immersive scenes with fantasy styles from reality, \ie, by giving text prompts, and automatically transferring textures of our living room with enchanting and meaningful details?

Over the past few years, enormous efforts have been paid in the field of scene 
stylization (or texture synthesis)~\cite{ARF, style_nerf, stylemesh, instruct-nerfpix2pix, SINE,text2tex,texture_paper}.
However, existing methods either only transfer low-level styles without semantically meaningful textures (\eg, imitating Van Gogh's paintings instead of generating recognizable visual elements~\cite{stylemesh, ARF}), or focus on texture editing~\cite{SINE, instruct-nerfpix2pix} on 3D objects with NeRF representation \cite{nerf} but struggle to generate high-fidelity textures for the whole space and achieve real-time rendering on HMD devices.
Very recently, with the advancements of diffusion-based generative methods (\eg, Stable Diffusion~\cite{rombach2022high}), it has become feasible to synthesize images based on text prompts with pleasant looking while maintaining the same scene structure by adding depth/edge conditions~\cite{controlnet,mou2023t2i}.
Nevertheless, since perspective image views only convey a partial appearance of the entire 3D scene, it's non-trivial to automatically project it to 3D scene geometries.
As a result, it usually requires skillful artists to run multiple generations and laboriously perform texture painting with 3D modeling software (\eg, Dream-Texture for Blender Addon~\cite{dreamtexture}).

In this paper, we propose a novel text-driven indoor scene texturing framework, which allows to generate meaningful and appealing mesh textures of real-world scenes based on text prompts, while preserving semantic consistency and spatial coherence (\eg, the furniture still looks like its own types but in different fashions, as shown in Fig. \ref{fig:teaser}).
Unlike the object texturing task~\cite{text2tex, texture_paper} that synthesize textures from multiple perspective views towards the object, for scene-level tasks, we should consider the panoramic semantics and consistency in a unified process to ensure a seamless texturing result (see Sec. \ref{ssec:exp_compare}).
To this end, we propose to texture scene meshes in a top-down manner, where we first generate an initial panoramic texture at the central viewpoint in a panoramic diffusion process and then propagate the panoramic texture to the rest of the regions.
Meanwhile, both the initial and the propagated textures will be baked into the resulting meshes through UV maps, which can be uploaded into a commodity-level HMD device for immersive VR applications (see the supplementary video for more details).

However, it is nontrivial to design such a scene-level mesh texturing framework in a top-down panoramic manner, since there are several challenges when texturing on unstructured and cluttered real-world scenes.
\textbf{1)}
To display sharp and visually comfortable content on HMD devices, the desired panoramic texture should be high-resolution, free of tiling seams to avoid the sense of spatial fragmentation, and spatially coherent following equirectangular projection (\eg, all the furniture and room structure such as floor and ceiling should be recognizable and not distorted).

To fulfill all the above demands, we employ a coarse-to-fine panoramic texture generation strategy, where we first generate a low-resolution panorama with a panoramic diffusion model to ensure proper panoramic scene structure, and then upscale it following equirectangular seam fixing to achieve seamless and high-resolution textures.
\textbf{2)}
Even with depth or edges as conditioning input~\cite{controlnet,mou2023t2i}, existing diffusion models cannot ensure adequate alignment between geometry and textures, and such misalignment would inevitably introduce noticeable texture projection artifacts (see Sec. \ref{ssec:exp_ablation} and Fig. \ref{fig:ablation_imitate_inpaint}).
To address this issue, we propose a novel dual texture alignment strategy, where the style-first textures and the alignment-first textures would be both generated and blended according to viewpoint depth changes.
In this way, we effectively mitigate the geometry-texture misalignment while preserving visually appealing generated styles.
\textbf{3)} 
Real-world reconstructed scenes often have intricate occlusions when observing from perspective views (\eg, narrow spaces such as the gap between the wall-mounted TV and the wall, or floor areas under the sofa, or thin structures like plant leaves or legs of furniture), making it challenging for viewpoint-based texture painting to effectively cover every aspect of the scene.
To this end, we design a holistic texture propagation pipeline.
Specifically, for regions free of occlusion from the new viewpoint, we employ diffusion-based~\cite{rombach2022high,controlnet} confidential texture inpainting.
Then, we leverage a coordinate-based implicit texture imitating network, which learns style mapping from real-world colors to stylized colors, and imitates textures for the rest of uncovered regions.
By cooperating inpainting and imitating techniques, our method smoothly propagates initial panoramic textures to the whole space while preserving spatial coherence.

We summarize the technical contribution as follows.
First, we propose a novel scene-level mesh texturing framework in a top-down panoramic manner, which allows users to generate engaging UV textures of real-world scene reconstructions based on text prompts.
Second, we develop a coarse-to-fine texture generation strategy to ensure the correct perspective and high resolution, and a dual texture alignment mechanism to alleviate geometrical misalignment without compromising style quality.
Moreover, to cope with the cluttered real-world geometries, we design a holistic texture propagation paradigm with inpainting and implicit imitating techniques, which smoothly paints the entire space with coherent textures.
Finally, extensive experiments on real-world datasets demonstrate that our method achieves significantly better scene-level mesh texturing quality than existing methods, which also brings immersive and impressive VR experiences when visualized on HMD devices.

\section{Related Works}

\noindent\textbf{Scene-Level Stylization.}
In the field of computer vision and graphics, neural network-based stylization has been studied for years.
Starting from Gatys \etal's work~\cite{gatys2016image}, early literature~\cite{chiu2020iterative,gatys2017controlling,johnson2016perceptual} mainly requires a style image as a reference, and optimize a perceptual loss or use a model to perform style transfer in 2D image domain~\cite{johnson2016perceptual,ulyanov2016texture,kolkin2019style,mechrez2018contextual}.
With the quick development of neural rendering techniques~\cite{nerf}, such style transfer pipeline has soon be deployed into 3D space domain~\cite{huang2022stylizednerf,ARF,chiang2022stylizing,chen2022upstnerf,fan2022unified}, which mainly inherit the perceptual loss paradigm to optimize the appearance of the view-dependent color field while freezing the density field.
To obtain meaningful stylization results, recent works also use larger-scale external data-driven priors (\eg, CLIP model~\cite{radford2021learning}) for style transfer (or editing)~\cite{instruct-nerfpix2pix,SINE}, which achieves stylized results that also follow human language prompts, but these works mainly cannot be scaled to large indoor scenes that allow immersive room touring.
However, during the rendering stage, NeRF-based methods typically require extensive computation due to network inference, which is not computational-friendly for all-in-one HMD devices. 
Hence, another line of works tries to directly stylize upon the scene meshes by hand-crafted annotation~\cite{hauptfleisch2020styleprop,fivser2018styleblit} or upon the point cloud~\cite{cao2020psnet}.
For example, Text2Scene~\cite{tan2019text2scene} optimizes scene-level mesh textures with differentiable local fields to satisfy users' prompts, but requires structured CAD scenes, which is not applicable for real-world scene reconstructions.
StyleMesh~\cite{stylemesh} proposes to operate neural style transfer on the parameterization of UV textures, which produces stylized mesh that can be feasibly rendered on standard graphics pipeline, but only transfer appearance up to global styles without strong semantic meaning (\eg, mimicking artists' stroke), which cannot ensure sufficient visual comfort when displayed in HMD devices.
Therefore, existing works for scene-level stylization either are not applicable for immersive indoor scene-scale scenarios with affordable computation on HMD devices~\cite{instruct-nerfpix2pix,SINE}, cannot support semantic meaningful style generation~\cite{huang2022stylizednerf,ARF,chiang2022stylizing,chen2022upstnerf,fan2022unified,stylemesh}, or require well-structured CAD model instead of real-world reconstruction~\cite{tan2019text2scene}.

\begin{figure*}[!t]
\centering
\includegraphics[width=1.0\linewidth, trim={0 0 0 0}, clip]{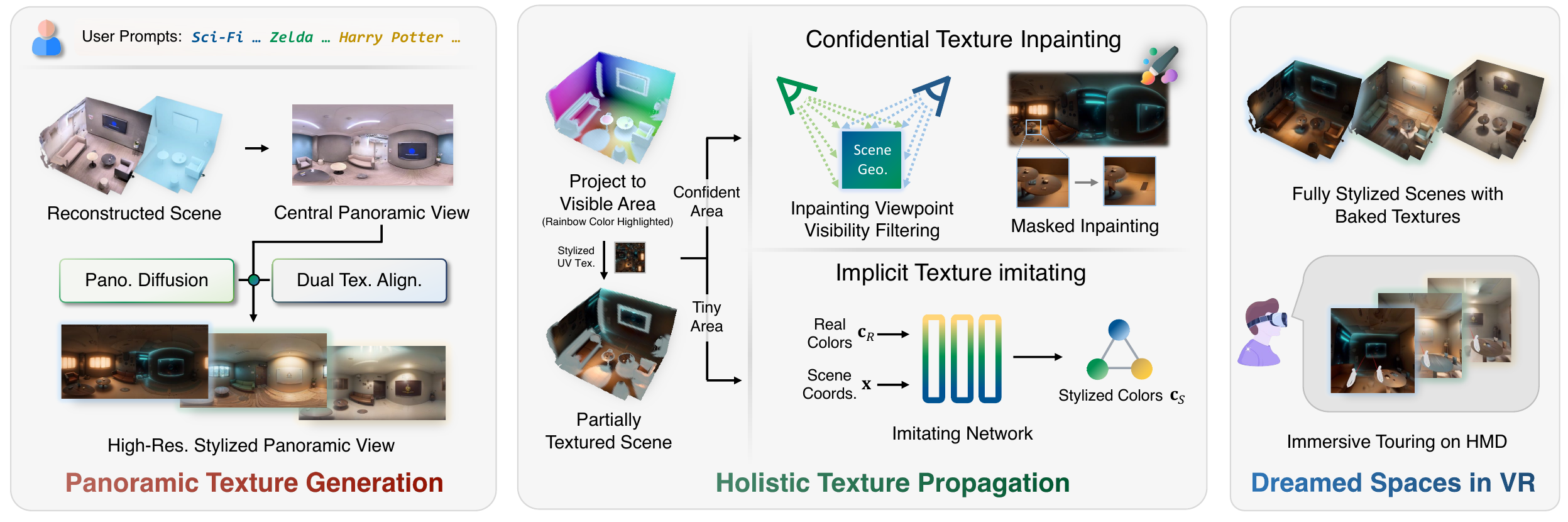}
\caption{
\textbf{Framework of DreamSpace.}
    Given a reconstructed real-world scene and users' text prompts, we first generate a high-resolution and geometrically aligned panoramic texture at the central viewpoint.
    Then, we propagate the textures into the rest regions with holistic texture propagation, where the confidential texture inpainting fills textures at the large confident areas and the implicit texture imitating predicts colors at the tiny areas.
    The resulting scene meshes with baked stylized UV textures can be uploaded into HMD devices for immersive VR touring.
    }
\label{fig:framework}
\end{figure*}

\noindent\textbf{Diffusion-based Mesh Texture Generation.}
Very recently, due to the emerging usage of large vision-language model in vision tasks, the generative methods~\cite{dhariwal2021diffusion,gao2022get3d,ho2022cascaded,nichol2021improved,saharia2022image,michel2022text2mesh,mohammad2022clip,akimoto2022diverse,wu2023ipo} have gained tremendous develop in the past few months.
Among them, diffusion-based generative models have attracted lots of attention in various modalities, such as high-resolution image generation~\cite{rombach2022high,SDXL}, human voice generation~\cite{liu2022diffsinger}, or even 3D model generation~\cite{poole2022dreamfusion,text2room}.
Notably, the open-source of Stable Diffusion also sparks a trend of AI-assisted creation throughout the whole community, which also derives a lot of following modules upon its pre-trained weights, such as injecting various controlling conditions~\cite{controlnet,mou2023t2i}, video generation~\cite{guo2023animatediff}, high-fidelity image inpainting~\cite{suvorov2022resolution} or even object texturing or mesh generations~\cite{latentnerf,texture_paper,text2tex}.
For example, Text2Room~\cite{text2room} uses Stable Diffusion to generate indoor 2D views, and lifted into 3D spaces with depth prediction and consecutive image inpainting, which enables to build up a novel indoor scene based on users' text prompts, but it struggles to produce clean textures or processes on a pre-captured scene reconstruction.
Therefore, for the mesh texture generation task with given targeting meshes, there are mainly two different pathways.
One is to use Score Distillation Sampling losses (SDS loss) from DreamFusion~\cite{poole2022dreamfusion}, which trains a generative NeRF by extracting supervisory signals from the denoising process of diffusion model upon the NeRF rendered views.
Inspired by DreamFusion, LatentNeRF~\cite{latentnerf} proposes to use SDS loss to paint textures on the exact mesh with the unwrapped UV texture map.
While the application of SDS over the mesh texturing task is technically plausible, it cannot unlock the full generative ability of the diffusion model, which results in much blurry rendering when compared with 2D domain image synthesis~\cite{rombach2022high,texture_paper,roomdreamer}.
Hence, another possible route is to first generate 2D textures~\cite{dreamtexture,text2tex,texture_paper} that align with 3D geometry using depth-aware conditioning techniques~\cite{controlnet,mou2023t2i}, and then project it into UV textures.
For example, the popular Blender addon Dream-Texture~\cite{dreamtexture} uses customized geometry node to render depth from interactive modeling views, and then projects the textures through the view frustum.
Nevertheless, since a single 2D viewpoint only reflects partial textures of a complete 3D model, Dream-Texture cannot correctly justify where to paint and simply projects textures through the entire mesh (\ie, back face with the same textures as the front face), which results in incorrect textures when viewing from 360$^{\circ}$ viewpoints.
To tackle the challenge of 2D-to-3D texturing ambiguity, TEXTure~\cite{texture_paper} and Text2Tex~\cite{text2tex} propose to synthesize multi-view textures from orbiting viewpoints aiming at the object center, and use depth-aware texture inpainting to fill the new unpainted areas while preserving consistent texture from the partially painted area.
However, such multi-view texturing pipeline assume the object can be fully observed without tiny / far-away structures or complex occlusions, which cannot be satisfied in real-world cluttered scenes.
Therefore, recent work MVDiffusion proposes to leverage 3D correspondence in an attention mechanism during the multi-view diffusing process, which achieves multi-view consistency to a certain degree but still cannot achieve satisfactory mesh texturing results (see Sec. \ref{ssec:exp_compare}).
Another concurrent work RoomDreamer tries to generate textures in cubemap format and also uses inpainting to fill the rest areas, but it still cannot ensure sufficient spatial coherence and also lacks proper ways to handle the unobserved regions (\eg, gap between the desk and the floor).
On the contrary, we propose to generate 360$^{circ}$ textures in the panoramic space with a coarse-to-fine panoramic diffusion process, and then propagate it into the rest region with inpainting and imitating, which both achieves texture synthesis with strong semantic meaning and takes into account the occlusion and tiny structures in real-world scene reconstruction.

\section{Method}

We introduce DreamSpace, a novel text-driven framework for generating semantically meaningful and spatial coherence scene textures for real-world indoor scenes.
As demonstrated in Fig. \ref{fig:framework}, we texture the scene in the panoramic space with a top-down fashion, where we first generate a stylized 360$^{\circ}$ view from the central viewpoint, and then propagate it to the entire scene.
To generate the high-resolution panoramic view with appropriate structure relationship and consistent semantic meaning, we design a coarse-to-fine panoramic texture generation process conditioned on reconstructed geometry and texture cues (Sec. \ref{ssec:method_texture_gen}), and a dual texture alignment strategy to alleviate texture misalignment to the geometry (Sec. \ref{ssec:method_texture_align}).
Once the initial stylized panoramic view is generated, we project textures to the visible area through UV maps, and then propagate it with confidential texture inpainting for visible areas at new viewpoints and implicit texture imitating for tiny areas, so as to obtain a fully stylized scene mesh.
Note that our method does not rely on volumetric rendering with any geometry approximation~\cite{nerf}. 
Therefore, the baked resulting mesh is exactly what you see during the generation, and is compatible with standard rendering pipelines, which then can be easily uploaded and experienced in all-in-one HMD devices without PC streaming.

\subsection{Panoramic Texture Generation}
\label{ssec:method_texture_gen}

\noindent\textbf{Generating in panoramic space.}
Different from previous object mesh texturing methods~\cite{texture_paper,text2tex,text2room} that repeatedly generates multiple perspective views towards object centers, we urge that the scene-level texture generating task should consider the full 360$^{\circ}$ view of the scene as a whole, \ie, generating in panoramic texture space (a.k.a. through equirectangular projection), rather than using multiple perspectives~\cite{texture_paper,text2tex} or cubemaps~\cite{roomdreamer} with perplexing viewpoint specific prompts (\eg, ``floor/ceiling in a single color'' when looking at the floor~\cite{text2room}).
To this end, given a user prompts $P$ and the reconstructed real-world scene (\ie, a textured scene mesh), our first attempt is to generate a vivid and high-resolution stylized panoramic view that observes the scene from a central viewpoint.
While it is plausible to use a depth-aware latent diffusion model (LDM)~\cite{rombach2022high,controlnet,mou2023t2i} to generate textures that fit to the observed scene depth, we find it still faced with several challenges.
First, existing generic or LoRA-fine-tuned LDMs cannot ensure accurate equirectangular projection, which results in distorted texture when projecting back to the mesh.
Second, the desired panoramic texture should be high-resolution (\eg, 2K resolution or more) and free of tiling seams to guarantee acceptable visual quality in immersive VR applications, which is also not directly feasible for texture generation methods.

\noindent\textbf{Coarse-to-fine conditioned generation.}
To handle the challenges above, we design a coarse-to-fine conditioned generation paradigm, where we first generate a low-resolution panoramic view with proper spatial structure, and then upscale it to the high resolution.
Specifically, we first train a panoramic diffusion model by fine-tuning generic LDM~\cite{rombach2022high} with carefully filtered equirectangular projected images (see the supplementary material for more details).
Next, for an input textured scene mesh, we render the panoramic colored image $I_{\text{P}}$ with distance map $D$ (\ie, distance from camera center $\textbf{c}$ to mesh surface) at the scene center, and feed them together with user's prompts $P$ to the fine-tuned LDM with multi-condition controls~\cite{controlnet} to obtain stylized image $\hat{I}_{S}$, as: 
\begin{equation}
    \hat{I}_{S} = F_\text{c}(P;D,\mathcal{E}(I_{\text{P}}))
\end{equation}
where $F_\text{c}$ is the LDM with multi-conditioning, $\mathcal{E}(I_{\text{P}})$ is the soft edgemap extracted with Su \etal's work~\cite{su2021pixel}.
During the inference, we adapt the asymmetric tiling strategy~\cite{asymmetric_tiling} by hijacking all the 2D convolutions of the UNet with horizontal circular padding for the last 60\% timestamps, so as to make sure the left and right side of the equirectangular image can be continuous (\eg, maintaining the wall and the furniture to keep the same tone and continuous patterns on both sides).
Then, we utilize tiled diffusion~\cite{multidiffusion} with a generic LDM to upscale the $\hat{I}_{S}$ into $\hat{I}_{SL}$, which produces 3 times larger panoramic images with extra rich details.

\noindent\textbf{Equirectangular seam fixing.}
During the upscaling stage, we find that the tiled upscaling strategy would inevitably break the equirectangular traits of the images (\ie, patterns become no longer tiling along the horizontal direction, and the top and lower part of the panoramic are not the correct stretching follows equirectangular projection), primary due to the reason that each processed tile is agnostic to the whole perspective knowledge.
Therefore, we also conduct inpainting on the top/down polar and left-right tiling side of the image.
Specifically, for the top/down polar, we unwrap the panorama to the upward and downward perspective view and inpaint the central disk area, and then warp it back.
For the horizontal tiling seam, we roll the half image along the x-axis and inpaint the middle part that covers both left-right sides of the panorama.
So far, we can obtain a high-resolution stylized panoramic image that satisfies equirectangular projection and also maintains semantic coherence.

\subsection{Dual Texture Alignment}
\label{ssec:method_texture_align}

\begin{figure}[!t]
\centering
\includegraphics[width=1.0\linewidth, trim={0 0 0 0}, clip]{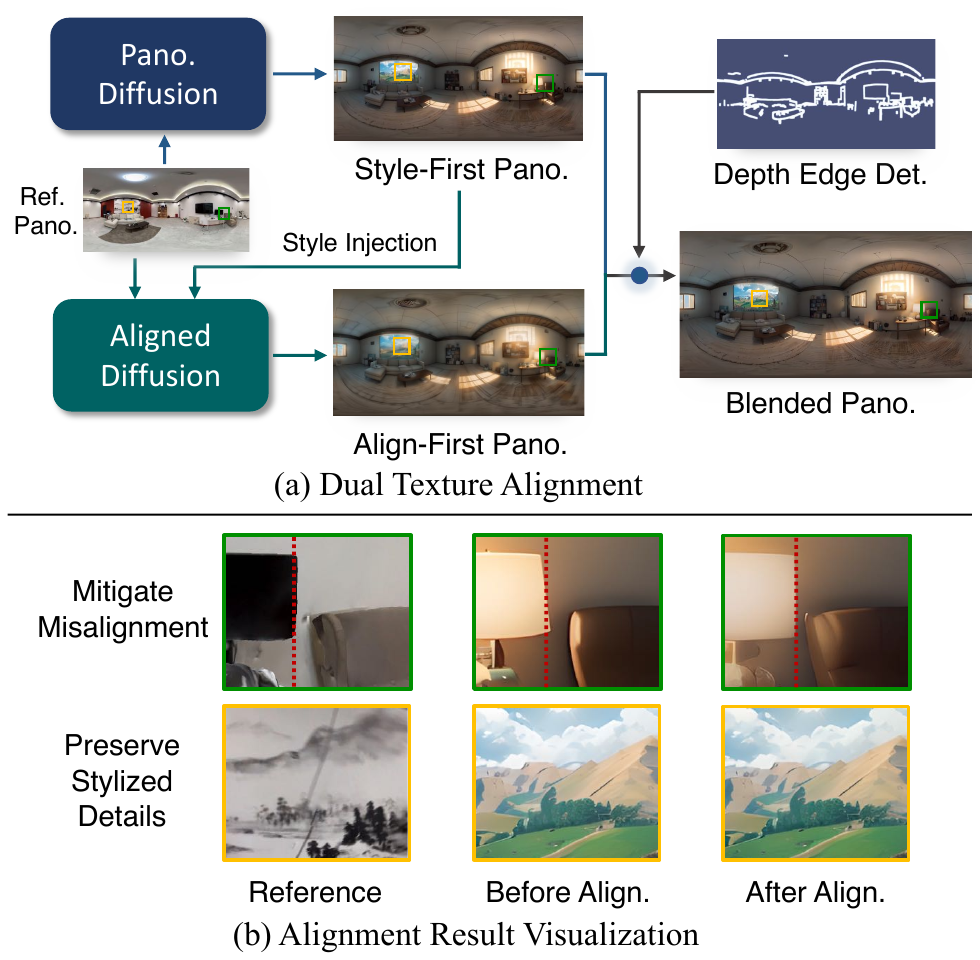}
\caption{
\textbf{Overview of dual texture alignment.}
To mitigate geometry-texture misalignment, we first synthesize style-first panorama and align-first panorama, and then blend these dual textures according to depth edge detection, which brings aligned panoramic textures while preserving visually appealing stylized details.
    }
\label{fig:method_align}
\end{figure}

\noindent\textbf{Dilemma of stylization and alignment.}
Although using depth or hedges as conditional control can effectively direct the LDM to produce somewhat consistent textures to the target mesh~\cite{controlnet,mou2023t2i,dreamtexture}, we find that in scene-level texturing tasks, such alignment is not sufficient since the geometry of the real-world scenes is generally much more complicated than single objects.
One plausible workaround might be directly denoising with moderate or small noises upon real image views (a.k.a. LDM's image-to-image mode with lower denoising strength).
However, due to the incomplete denoising process, such a method would generally result in blurry images or unsatisfactory styles.
Therefore, we are faced with a dilemma that the visually appealing viewpoint stylization and perfect geometric alignment cannot be achieved together at one time.

\noindent\textbf{Alignment with dual texture blending.}
To solve the dilemma, we propose to break the stylized panoramic texture generation in a dual process, and then fuse the dual textures in a geometry-aware manner, as demonstrated in Fig. \ref{fig:method_align}.
For brevity, we named these dual textures style-first panorama and align-first panorama (see the middle part of Fig. \ref{fig:method_align} (a)), where the first one is synthesized in a way as introduced in Sec. \ref{ssec:method_texture_gen} which ensures high-quality styles, and the second one is synthesized with a customized aligned diffusion process that tends to align the original scene more strictly while maintaining a similar style.
Specifically, for generating align-first panorama $\hat{I}_{A}$, we start by denoising on the real-world reference panorama but utilize multi-control techniques~\cite{controlnet}
, as:
\begin{equation}
    \hat{I}_{A} = F_\text{c}(P;\mathcal{C}(I_{\text{P}}),\mathcal{T}(I_{{S}})),
\end{equation}
where $\mathcal{C}(I_{\text{P}})$ is the canny edge control that enforces alignment, and $\mathcal{T}(I_{{S}})$ is the tile control~\cite{controlnet} that injects styles from the style-first panorama.
To make the same size as $\hat{I}_{SL}$, we upscale the $\hat{I}_{A}$ into $\hat{I}_{AL}$ with Wang's work~\cite{wang2021realesrgan}, which empirically would not introduce noticeable tiling seams.
Note that we do not need this panorama to be perfectly stylized (which in practice is noticeably blurry than the style-first one, as shown in Fig. \ref{fig:method_align} (a)).
Then, we determine the pixel areas for blending the align-first panorama with the style-first panorama.
We observe that the misalignment issue generally happens where the scene depth changes evidently.  
Hence, we simply generate the blending mask by detecting depth edges from the panoramic depth map following the dilation and blurring operations, and then blend these dual textures with masked Poisson image editing~\cite{perez2023poisson} (a.k.a. seamless cloning with the align-first panorama as the source and style-first panorama as the target).
In this way, we can successfully mitigate the geometry-texture misalignment while maintaining the desired stylized details untouched (see Fig. \ref{fig:method_align}(b), where the edge of the black monitor and sofa are much better aligned, while the stylized posters on the wall keep unchanged).

\subsection{Holistic Texture Propagation}
\label{ssec:method_texture_prop}

\noindent\textbf{Panoramic texture projection through UV maps.}
Once the initial stylized panoramic view is synthesized, we project it to the visible areas through UV maps in the panoramic space, as illustrated in Fig. \ref{fig:framework} (the left column of the holistic texture projection).
In practice, we first obtain scene coordinates $\textbf{x}$ (3D position) for valid pixels $\textbf{p}$ in the corresponding UV map, as:
\begin{equation}
    \textbf{x} = \text{Interp}(\text{MapTex}(\text{TexCoord}(\textbf{p}),\{T\})),
    \label{eq:rast_scene_coord}
\end{equation}
where $\text{TexCoord}(\textbf{p})$ is the texture coordinate of each $\textbf{p}$, $\{T\}$ is the mesh triangles, $\text{MapTex}(\cdot)$ maps the texture coordinate into triangle vertices with barycentric weights, and each $\textbf{x}$ is barycentric interpolated from the triangles' vertices.
Next, for each $\textbf{x}$, we compute ray directions from the observing camera center $\textbf{c}$, and map the direction $\mathbf{d} = {\mathbf{c} - \mathbf{x}} / {\|\mathbf{c} - \mathbf{x}\|}$ to the panoramic space through equirectangular projection.
Then, for each $\textbf{x}$, we compare its observing distance to the rendered scene depth and determine if the corresponding UV pixel $\textbf{p}$ is visible from the viewpoint with a distance threshold $\epsilon=0.01$.
We go through all the UV pixels with the visibility test and form an initial visibility mask $M_\text{init\_vis}$ on the UV space, as:
\begin{equation}
    M_\text{init\_vis}(\textbf{p})=
    \begin{cases}
        1, & \text{if } \|\textbf{p} - \textbf{x}\| < \epsilon \\
        0, & \text{otherwise}.
    \end{cases}
    \label{eq:visibility}
\end{equation}
Finally, we assign stylized panoramic colors to the UV spaces according to the initial visibility mask $M_\text{init\_vis}$ and corresponding ray directions $\mathbf{d}$, which produces the partially textured scene (see the middle part of Fig. \ref{fig:framework}).

\noindent\textbf{Separated strategies for confidential and tiny areas.}
By projecting initial textures to the scene, the main impression of the styled space has been already shaped, while there are still some uncovered areas that need to be filled (\eg, the gray region at the partially textured mesh in Fig. \ref{fig:framework}).
Previous methods that use LDM for object mesh texturing~\cite{texture_paper,text2tex} mainly rely on inpainting with various area selection and masking methods (\eg, maintaining a trimap by TEXTure~\cite{texture_paper}), which aim to cover the entire mesh surface as complete as possible.
However, for real-world scene texturing with cluttered geometries, solely relies on automatic inpainting cannot ensure proper texturing for thin structures (\eg, leaves and furniture legs) or severely occluded areas (\eg, floor under the sofa or gaps between wall-mounted TV and the wall) that cannot be observed from normal camera positions. 
Besides, duplicated inpainting on the same area of the mesh surface would also result in blurry appearance or artifacts due to the inconsistency nature of LDM's inpainting result (as demonstrated in Sec. \ref{ssec:exp_compare}).
Therefore, we propose separate strategies for areas with different visibility.
Instead of conducting inpainting multiple times, we only inpaint at the confidential areas (\ie, areas that is definitely free of occlusion) in very few viewpoints (\eg, only two in our experiments) and then leverage a novel implicit texture imitating network to smoothly fill the rest of areas with plausible appearance.

\begin{figure}[!t]
\centering
\includegraphics[width=1.0\linewidth, trim={0 0 0 0}, clip]{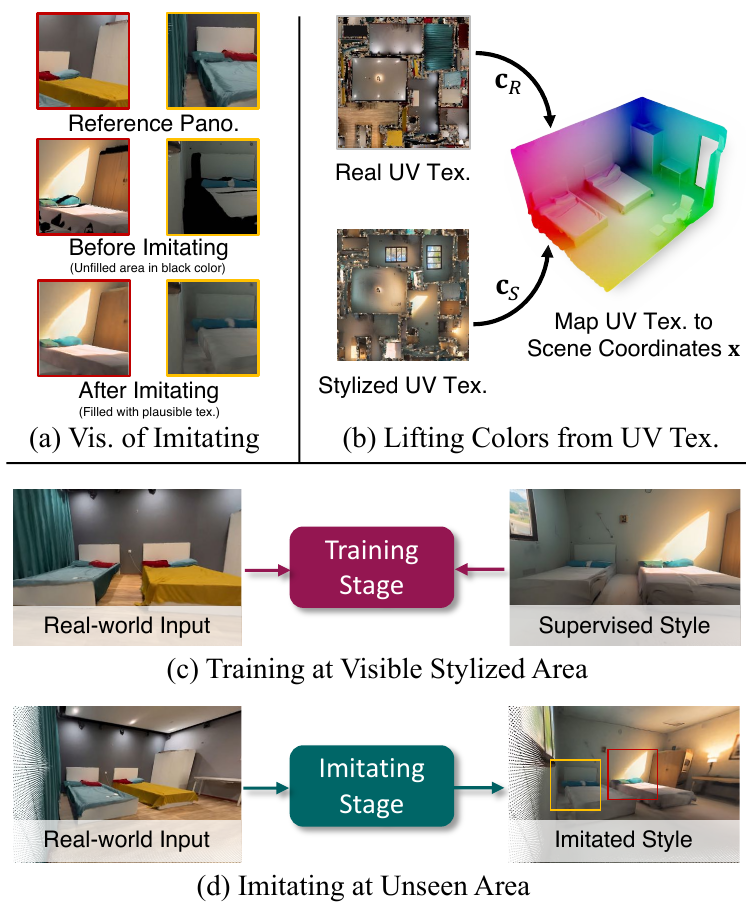}
\caption{
    \textbf{Overview of implicit texture imitating.}
    We first lift colors from UV textures according to UV pixels' scene coordinates.
    Then, during the training stage, we train an implicit texture imitating network from visible stylized areas using lifted real-world/stylized colors and coordinates.
    During the imitating stage, we feed the real-world color and coordinates into the network to imitate plausible textures in unseen areas.
    }
\label{fig:method_imitate}
\end{figure}

\noindent\textbf{Confidential Texture Inpainting.}
Given a partially textured mesh, we first perform confidential texture inpainting in the panoramic space as demonstrated in the middle part of Fig. \ref{fig:framework}.
During this procedure, we do not aim to fill every aspect of the space, but only cover the confidential areas that are totally free of occlusion when observing from new viewpoints, where the viewpoint can be selected by SfM poses with farthest point sampling or interactive user selection.
To begin with, for each viewpoint, we first determine the panoramic inpainting mask $M_\text{inp}$ from the new camera poses.
Practically, we reuse the UV-space initial visibility mask $M_\text{init\_vis}$ by regarding it as the UV texture, and render the panoramic image on the current viewpoint, and then perform dilation and blurring to the image to obtain the $M_\text{inp}$.
We then leverage depth-aware inpainting LDM~\cite{rombach2022high,controlnet} $F_\text{inp}$ to synthesis masked areas, as:
\begin{equation}
    \hat{I}_\text{inp} = F_\text{inp}(P,\hat{I}_\text{M};D,M_\text{inp}),
\end{equation}
where $\hat{I}_\text{M}$ is the rendered panoramic image with partially textured mesh, $\hat{I}_\text{inp}$ is the inpainting output image.
Note that the inpainting results $\hat{I}_\text{inp}$ will not be fully projected into the stylized UV texture, but only retain confidential areas by UV space masked filtering.
More specifically, we design three UV-space mask filters that ensure a confidential texture projection.
First, we filter inpainting areas with abrupt depth changes using a depth edge filtering mask $M_\text{dep\_edge}$, which can be constructed by assigning the UV mask with panoramic depth edge detection as introduced in Sec. \ref{ssec:method_texture_align}.
Second, we consider the surface normal and distances by rejecting small grazing viewing angles ($10^{\circ}$) or too far surface points (distance larger than $2.5$ meters) to form a safe viewing mask $M_\text{safe\_view}$, which is constructed by calculating barycentric interpolated normal vectors from vertex normal for each valid UV pixel along with the scene coordinates.
Third, we perform visibility test on the inpainting views with a similar formulation as Eq.~(\ref{eq:visibility}), which constructs the inpainting visibility mask $M_\text{inp\_vis}$.
We combine all the above masks to achieve a confidential texture projecting areas in UV space, as:
\begin{equation}
    M_\text{conf} = M_\text{dep\_edge} \cap M_\text{safe\_view} \cap M_\text{inp\_vis},
\end{equation}
where $M_\text{conf}$ is the combined confidential mask.
Note that all the masks are constructed in UV space instead of a certain camera perspective or panoramic view, which avoids the influence of viewpoint-specific occlusion.
We assign inpainting panoramic texture into the stylized UV texture with the mask $M_\text{conf}$, which further fills the partially stylized scenes with more textures.

\noindent\textbf{Implicit Texture Imitating.}
To complement the unobserved or unpainted areas for scene-level mesh texturing, we design a novel implicit texture imitating mechanism.
As demonstrated in Fig. \ref{fig:method_imitate}, the goal of the texture imitating is to learn the style mapping from the partially stylized scenes, and then smoothly predict plausible texture for unseen areas.
In practice, we first lift real-world colors $\mathbf{C}_R$ and stylized colors $\mathbf{C}_S$ from the corresponding UV textures into the scene coordinates $\mathbf{x}$ (see Eq.~(\ref{eq:rast_scene_coord}) and Fig. \ref{fig:method_imitate} (b)).
During the training stage (see Fig. \ref{fig:method_imitate} (c)), we learn an implicit imitating network $F_\text{Imit}$ (\ie, a coordinate-based MLP), which gives the input as scene coordinate $\mathbf{x}$ and real-world colors $\mathbf{C}_R$ from the partially textured scenes, and is supervised by existing visible stylized colors $\mathbf{C}_S$ with L2 loss, as:
\begin{equation}
    \mathcal{L}_\text{imit} = \lVert \hat{\mathbf{C}}_S - \mathbf{C}_S \rVert_2\;, \text{where} \;\;\mathbf{C}_S=F_\text{Imit}(\gamma(\textbf{x}),\mathbf{C}_R),
\end{equation}
where $\gamma(\cdot)$ is the positional encoding~\cite{nerf}, and $\hat{\mathbf{C}}_S$ is the predicted imitating color.
Then, during the imitating stage (see Fig. \ref{fig:method_imitate} (d)), we feed the network with all the valid UV pixels' scene coordinates $\mathbf{x}$ and real-world colors $\mathbf{C}_R$ to predict the imitated colors $\hat{\mathbf{C}}_S$.
As visualized in Fig. \ref{fig:method_imitate} (a), the uncovered areas in the stylized scene can be smoothly filled after the imitating while also preserving spatial coherence (\eg, the pillows and the bedsheet are faithfully predicted as blue and white textures).
Finally, we fuse the imitated colors into the partially textured meshes through the accumulated visibility mask $M_\text{accu}$ (by combing $M_\text{init\_vis}$ and all the $M_\text{inp\_vis}$), which produces the fully stylized scenes with baked textures, as demonstrated in the right part of Fig. \ref{fig:framework}.

\begin{figure*}[!t]
\centering
\includegraphics[width=1.0\linewidth, trim={0 0 0 0}, clip]{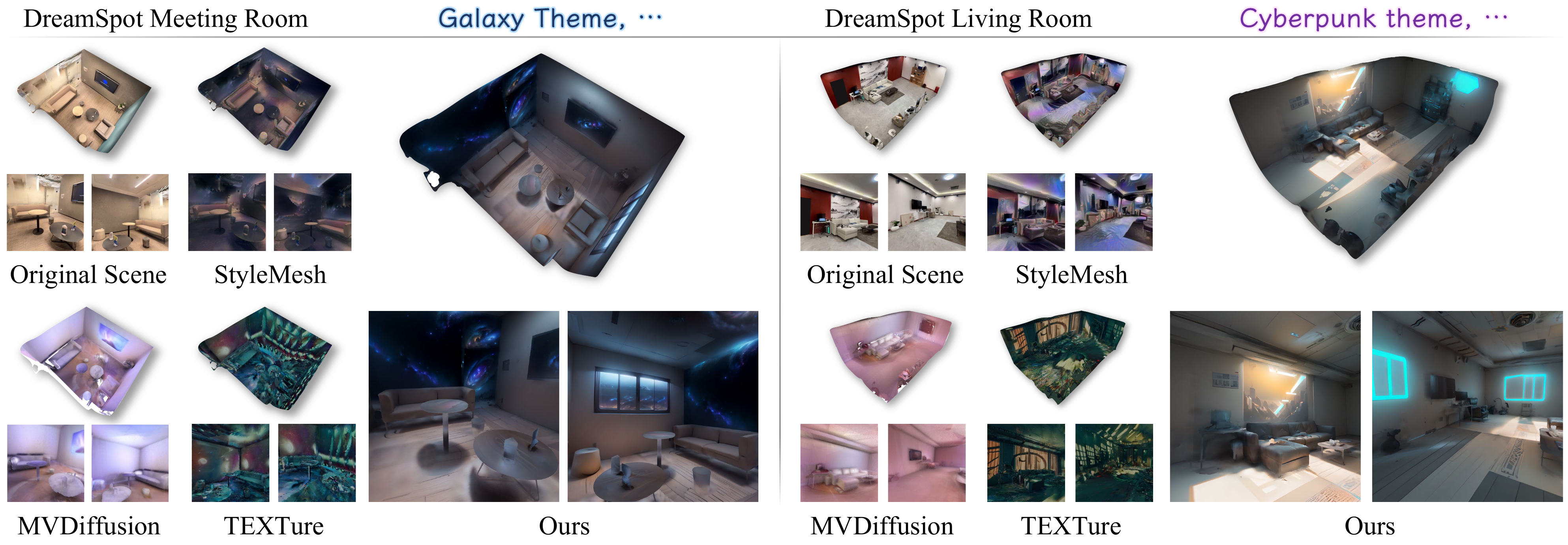}
\caption{
    We compare our scene-level mesh texturing with StyleMesh~\cite{stylemesh}, MVDiffusion~\cite{mvdiffusion} and TEXTure~\cite{texture_paper} on our captured DreamSpot dataset, where the figures include the overview of textured scene meshes and the corresponding rendered views.
    }
\label{fig:compare_our_scene}
\end{figure*}

\begin{figure*}[!t]
\centering
\includegraphics[width=1.0\linewidth, trim={0 0 0 0}, clip]{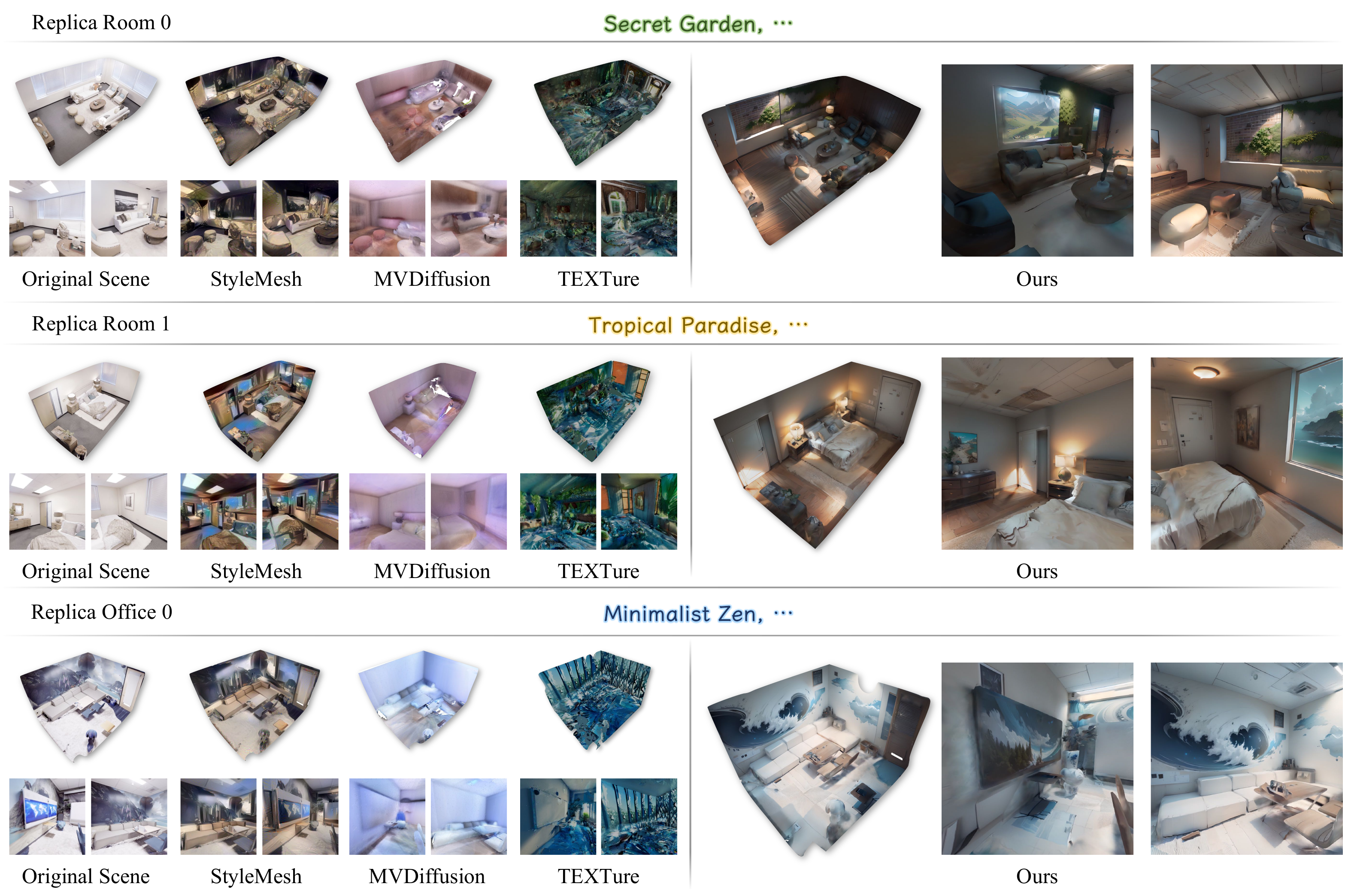}
\caption{
    We compare our scene-level mesh texturing with StyleMesh~\cite{stylemesh}, MVDiffusion~\cite{mvdiffusion} and TEXTure~\cite{texture_paper} on the Replica dataset, where the figures include the overview of textured scene meshes and the corresponding rendered views.
    }
\label{fig:compare_replica}
\end{figure*}

\section{Experiments}

In this section, we first compare our framework with existing methods on the generative scene-level mesh texturing task (Sec. \ref{ssec:exp_compare}) on real-world indoor scene datasets.
Next, we analyze the necessity of panoramic space texture synthesis by comparing it with the cubemap space (Sec. \ref{ssec:exp_pano_or_cubemap}).
Then, we perform ablation studies on the design of our texturing framework (Sec. \ref{ssec:exp_ablation}).
Finally, we build up an immersive VR application by uploading fully textured scenes into the HMD devices (Sec. \ref{ssec:exp_vr}).

\subsection{Datasets}
\label{ssec:exp_dataset}

\noindent\textbf{DreamSpot Dataset.} To demonstrate the applicability in real-world indoor scenes, we create a new dataset named DreamSpot, which contains three scenes that cover several typical scenarios in daily lives (\ie, meeting room, living room, and bedroom, where the first two are used for comparison).
Specifically, we use an iPhone to capture RGB images of the room and then use out-of-box SfM~\cite{schonberger2016structure} with MonoSDF~\cite{yu2022monosdf} for geometric reconstruction, and utilize texture mapping~\cite{openmvs2020open} to obtain scene meshes with real-world UV textures.

\noindent\textbf{Replica Dataset.}
We also use three real-world scenes from the Replica dataset~\cite{replica19arxiv} to evaluate our method, \ie, Room 0, Room 1, and Office 0.
Since the original Replica dataset uses a customized shader for HDR rendering, which is not directly compatible with textured mesh-based pipelines such as our method and StyleMesh~\cite{stylemesh}.
Hence, we first pre-process these scenes by baking the appearance into unwrapped UV textures with Blender.

\subsection{Comparison on Generative Mesh Texturing}
\label{ssec:exp_compare}

\noindent\textbf{Experiment setting.}
We first evaluate our method by comparing it with SOTA mesh texturing (or stylization) works on the scene-level meshes both quantitatively and qualitatively.
Specifically, given a reconstructed textured scene mesh and user-defined text prompts (\eg, ``galaxy themes'', or ``secret garden''), our task is to synthesize textures that fit the scene geometry while following the semantic meaning of the prompts.
We choose the UV texture stylization method (StyleMesh~\cite{stylemesh}), multi-view consistent 2D diffusion model (MVDiffusion~\cite{mvdiffusion}), and LDM-based depth-aware mesh texturing method (TEXTure~\cite{texture_paper}) as competitors.
Note that not all methods can directly process on meshes or leverage existing textures, \ie, StyleMesh and our method use real-world textures and geometry as input, while TEXTure and MVDiffusion can only use pure geometry or 3D correspondence as guidance, and MVDiffusion also uses TSDF fusion to fuse generated images into colored meshes.
For StyleMesh, since it uses perceptual loss for style transfer and requires a reference style image, we additionally use LDM~\cite{rombach2022high} with text prompts to generate a style image as its input.
During the texturing process, all the other methods perform optimization or generation in perspective views, while our method uses panoramic views.
Therefore, to make a fair comparison, we manually designed a perspective camera scanning trajectory for each scene with the best effort to cover the whole space while avoiding being too close to the mesh surface.
Once the mesh texturing is finished, we render the textured mesh into multiple perspective views with OpenGL, which will be used for metric comparisons and user study.

\begin{table}[t!]
\centering
\resizebox{1.0\linewidth}{!}{
\tabcolsep 2pt
\begin{tabular}{lcccccc}
\toprule
\multicolumn{1}{c}{\multirow{2}{*}{Methods}} & \multicolumn{2}{c}{Quantitative Metrics} & \multicolumn{2}{c}{User Study} \\ \cmidrule(lr){2-3} \cmidrule(lr){4-5}  
\multicolumn{1}{c}{} & \multicolumn{1}{l}{CLIP Score $\uparrow$} & \multicolumn{1}{l}{Aesthetic $\uparrow$} & \multicolumn{1}{l}{Correctness $\uparrow$} & \multicolumn{1}{l}{Quality $\uparrow$} \\ \hline
StyleMesh~\cite{stylemesh} & 0.184 & 4.812 & 2.68 & 2.76 \\
MVDiffusion~\cite{mvdiffusion} & 0.174 & 4.263 & 1.37 & 1.49 \\
TEXTure~\cite{texture_paper} & 0.187 & 5.265 & 2.57 & 2.20 \\
Ours & \textbf{0.214} & \textbf{5.771} & \textbf{3.38} & \textbf{3.55} \\
\bottomrule
\end{tabular}
}
\newline
\caption{
We perform quantitative evaluation and user studies on the rendered views of textured mesh for StyleMesh~\cite{stylemesh}, MVDiffusion~\cite{mvdiffusion}, TEXTure~\cite{texture_paper} and our method.  
}
\label{tab:compare}
\end{table}

\begin{figure}[!t]
\centering
\includegraphics[width=1.0\linewidth, trim={0 0 0 0}, clip]{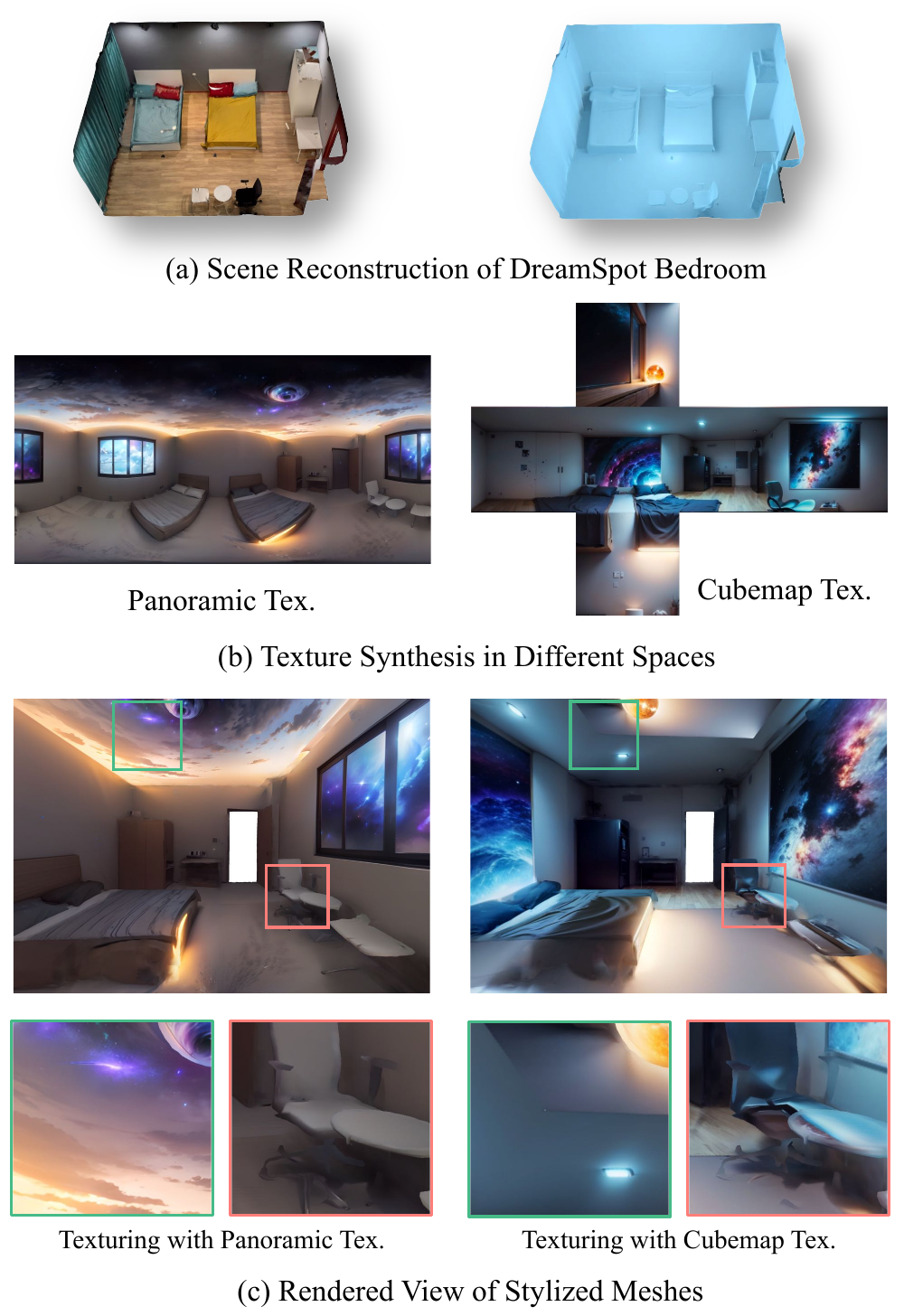}
\caption{
    We compare mesh texturing with textures generated from different spaces (\ie, panoramic texture or cubemap texture).
    }
\label{fig:compare_pano_cubemap}
\end{figure}

\begin{figure*}[!t]
\centering
\vspace{-1.5em}
\includegraphics[width=1.0\linewidth, trim={0 0 0 0}, clip]{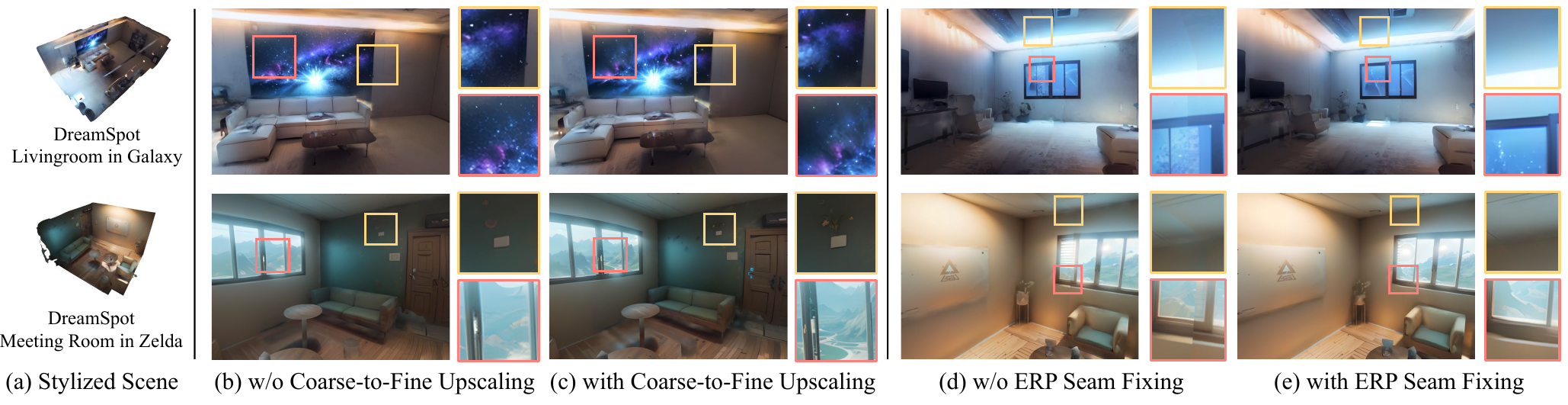}
\caption{
    We perform ablation studies of the coarse-to-fine strategy during the panoramic texture generation, including the coarse-to-fine upscaling and equirectangular seam fixing.
    }
\label{fig:ablation_coarse_to_fine}
\end{figure*}

\noindent\textbf{Quantitative comparison.}
For quantitative comparison, we use CLIP Score~\cite{radford2021learning} to measure the matching degree between rendered views and the given text prompts.
Besides, we also use aesthetic scoring introduced by LAION~\cite{aesthetic_score} to measure the aesthetic quality of the generated images, since it has been proven to be more authentic than FID for recent diffusion-based generative methods~\cite{SDXL}.
As presented in Fig. \ref{tab:compare}, our method consistently achieves the highest scores in both metrics, which demonstrates that our synthesized texture follows the given text prompts better and also maintains high quality when rendered from perspective views. 

\noindent\textbf{Qualitative comparison.}
We visualize the qualitative comparison results in Fig. \ref{fig:compare_our_scene} and Fig. \ref{fig:compare_replica}, where we both exhibit the overview of the fully textured meshes and the corresponding perspective mesh rendering views.
For StyleMesh, since it utilizes VGG perceptual loss~\cite{johnson2016perceptual} for UV texture style transfer without high-level semantic priors such as CLIP~\cite{radford2021learning}, it generally cannot synthesize novel and meaningful textures and behaves more like mimicking strokes and color tones of the given style image.
For example, in the ``galaxy theme'' of the meeting room (see Fig. \ref{fig:compare_our_scene}), StyleMesh mainly turns the environment into dark galaxy tones while failing to generate rich galaxy textures.
For MVDiffusion, though it leverages corresponding attention module to preserve multi-view consistency by extracting 3D correspondence from camera poses and scene depths, we find the resulting synthesized images cannot fulfill the requirement of scene-level texturing task due to the insufficient consistency, which results in blurry appearance in most of the cases (\eg, for both cases in Fig. \ref{fig:compare_our_scene}, the boundary of stylized television is much blurrier than ours).
For TEXTure, because its repetitive inpainting strategy is mainly designed for object meshes, we find it struggles to generate satisfactory textures when conducting on scene-level meshes (\eg, in Fig. \ref{fig:compare_replica} Replica Office 0, it produces repetitive artifacts on the walls) and also fails to project textures into scenes with cluttered geometry (\eg, pieces of unpainted areas in Fig. \ref{fig:compare_replica} Reolica Room 0).
To avoid potential visual discomfort, we have slightly dimmed the results of TEXTure in Fig. \ref{fig:compare_our_scene} and Fig. \ref{fig:compare_replica}.
From the analysis above, we believe that relying on perspective view for generating indoor scene textures is fairly difficult to obtain spatial coherent and consistent results, and also struggles to cover every visible area of real-world complex scenes.
By contrast, our method uses panoramic scene texturing, which not only preserves semantic meaning (\eg, furniture still looks like furniture, but in fantasy styles, and the generated floor texture is free of excessive details or severe artifacts), but also creates novel and enchanting textures by faithfully projecting generated textures into the meshes (\eg, galaxy on the floor in Fig. \ref{fig:compare_our_scene} meeting room galaxy theme, vibrant grass decorations in Fig. \ref{fig:compare_replica} Room 0 ``secret garden'', and the impressive landscape poster in Fig. \ref{fig:compare_replica} Room 1 ``tropical paradise''), while also properly fills unseen spaces (\eg, areas under the chair in Fig. \ref{fig:compare_replica} Office 0 ``minimalist zen'') thanks to texture propagating techniques.

\noindent\textbf{User study.}
We also conducted a user study to compare our method with others on the generated mesh textures of the DreamSpot and Replica datasets.
Specifically, we ask 20 users to sort the rendered views from textured meshes generated by methods in two aspects, \ie, image-text matching correctness and the perceptual quality, and assign the scores by their ranking (\ie, with a score of 4 for the ordered best one and a score of 1 for the last one).
As reported in Fig. \ref{tab:compare}, we achieve the most preferences among all the methods by a large margin, which highlights the impressive visual quality and image-text matching degree of our method.

\subsection{Panoramic Texture vs. Cubemap Texture}
\label{ssec:exp_pano_or_cubemap}

We suggest that, to pursue global consistency and spatial coherent for the scene-level mesh texturing task with the LDM diffusion process, the texture should be first synthesized in a panoramic space with equirectangular projection, rather than using multi-view fashion (\eg, as shown in Sec. \ref{ssec:exp_compare}) or cubemap spaces (\eg, RoomDreamer~\cite{roomdreamer}).
To prove this, we also compare our panoramic texturing pipeline with a cubemap-based pipeline, where the cubemap is directly generated by depth-aware LDM following Song \etal's work~\cite{roomdreamer}.
As demonstrated in Fig. \ref{fig:compare_pano_cubemap}, due to the discontinuity and unclear spatial semantic meaning, cubemap textures tend to produce excessive details on top faces, and also fail to make a smooth content transition on disconnected edges (see Fig. \ref{fig:compare_pano_cubemap} (b)), which results in spurious textures on the rooftop and mixed textures on the chair (see Fig. \ref{fig:compare_pano_cubemap} (c)).
By contrast, generating textures in panoramic space like ours not only achieves better spatial structural meaning (\ie, let the fine-tuned LDM know that the upper image area is the ceiling and the bottom area is the floor), but also ensures spatial continuity and coherence (\eg, semantic meaningful galaxy ceiling and white chairs with clean textures in Fig. \ref{fig:compare_pano_cubemap}).

\begin{figure}[!t]
\centering
\includegraphics[width=1.0\linewidth, trim={0 0 0 0}, clip]{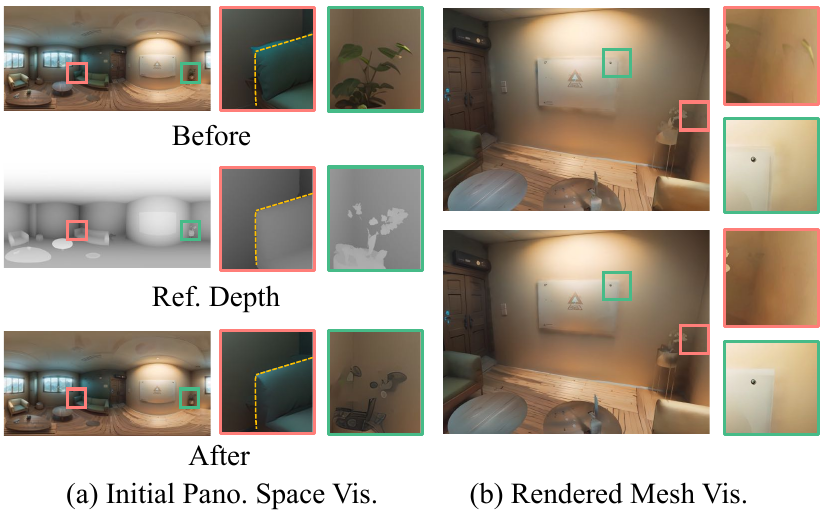}
\caption{
We inspect the efficacy of dual texture alignment on the initial panoramic space and rendered mesh.
    }
\label{fig:ablation_dual_align}
\end{figure}

\subsection{Ablation Studies}
\label{ssec:exp_ablation}

\noindent\textbf{Coarse-to-fine generation.}
We first analyze the coarse-to-fine strategy in panoramic texture generation (Sec. \ref{ssec:method_texture_gen}).
Specifically, we ablate the coarse-to-fine upscaling and equirectangular seam fixing for the initial panoramic texture generation.
As shown in Fig. \ref{fig:ablation_coarse_to_fine} (b) and (c), by enabling the coarse-to-fine upscaling technique, we can obtain textures with richer details (\eg, much cleaner galaxy-style poster, seeing clearer winding landscape path from the window), which is essential for satisfactory immersive VR experience as it amplifies the details of the scene.
By employing equirectangular seam fixing (see Fig. \ref{fig:ablation_coarse_to_fine} (d) and (e)), we can significantly remove tiling seams on the projected mesh textures (\eg, seams on the window and the roof are gently removed), which ensures the spatial consistency for the synthesized panoramic texture.

\noindent\textbf{Dual texture alignment.}
We then study the necessity of the dual texture alignment strategy (Sec. \ref{ssec:method_texture_align}).
To clearly demonstrate the efficacy, we both visualize the panoramic space alignment and the resulting meshes in Fig. \ref{fig:ablation_dual_align}.
It is clear that LDM tends to produce textures where the boundary of the object cannot be aligned to the real-world geometry (\eg, the highlighted contour of the green sofa, and the leaves of a potted plant in Fig. \ref{fig:ablation_dual_align} (a)), while dual texture alignment would mitigate such misalignment at the panoramic space.
After projecting textures to meshes following Sec. \ref{ssec:method_texture_prop} with carefully visibility test, we still observe the artifacts by misalignment (\eg, dirty textured walls caused by erroneously projecting leaves' textures on the wall in the first row of Fig. \ref{fig:ablation_dual_align}).
By introducing dual texture alignment for panoramic textures, we further alleviate the misaligned artifacts caused by texture projection (\eg, clean textured walls in the second row of Fig. \ref{fig:ablation_dual_align}).

\begin{figure}[!t]
\centering
\includegraphics[width=1.0\linewidth, trim={0 0 0 0}, clip]{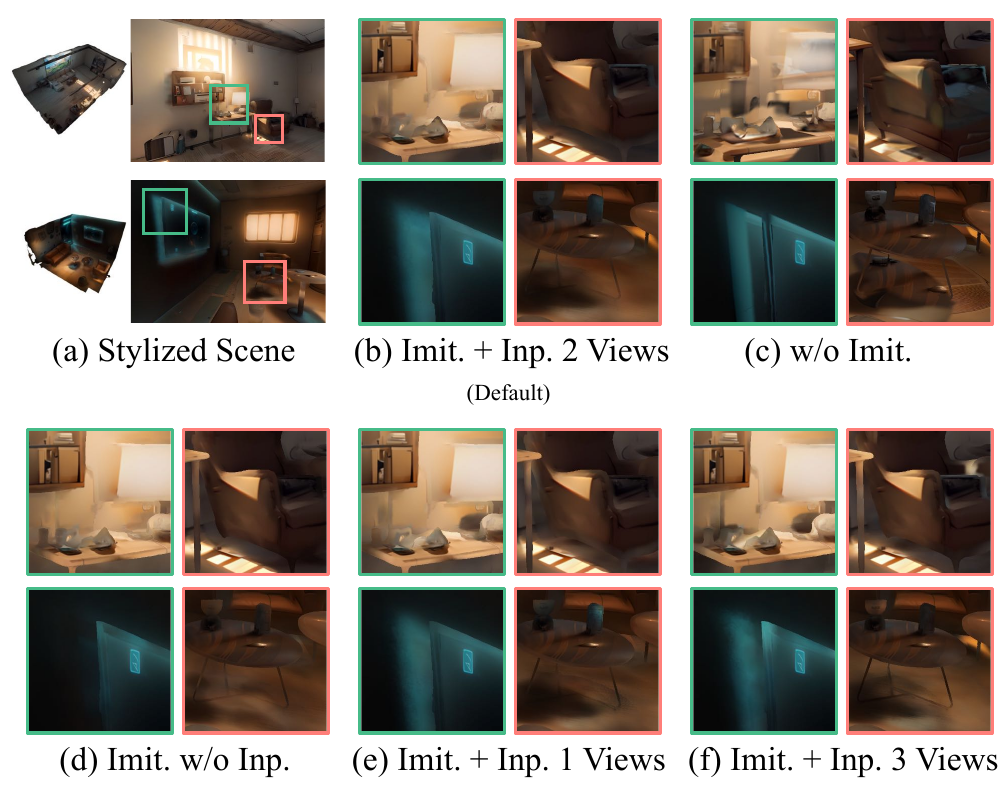}
\caption{
We analyze the effectiveness of imitating and inpainting in holistic texture propagation.
    }
\label{fig:ablation_imitate_inpaint}
\end{figure}

\noindent\textbf{Texture propagation with inpainting and imitating.}
We also inspect the necessity of the texture inpainting and imitating techniques (Sec. \ref{ssec:method_texture_prop}) for panoramic texture projection in Fig. \ref{fig:ablation_imitate_inpaint}.
By default, we enable texture imitating with two viewpoint inpainting (see Fig. \ref{fig:ablation_imitate_inpaint} (b)).
To ablate the texture imitating, we use a see-through texture projecting similar to Dream-Texture~\cite{dreamtexture} to avoid texturing vacancy, where all the valid UV pixels would be assigned to a color through equirectangular projection.
As shown in Fig. \ref{fig:ablation_imitate_inpaint} (c), the texture projection without imitating would inevitably introduce erroneous texturing results, \eg, much chaotic appearance of the desk and duplicated round table on the floor in the first row of Fig. \ref{fig:ablation_imitate_inpaint} (c).
When ablating texture inpainting techniques, the framework loses knowledge of what the occluded area should look like and only guesses the occluded appearance with texture imitating.
As shown in Fig. \ref{fig:ablation_imitate_inpaint} (d), our method still achieves plausible texturing results without noticeable artifacts, but might lose some semantic meaningful content such as the blue glow at the back of the monitor (the last row of Fig. \ref{fig:ablation_imitate_inpaint} (d)).
By enabling the inpainting and imitating together, we can achieve texturing results with both clean textures at cluttered geometry (\eg, the first row of Fig. \ref{fig:ablation_imitate_inpaint} (e)) and novel content at inpainted areas (\eg, the fancy blue glow of the monitor at the last row of Fig. \ref{fig:ablation_imitate_inpaint} (d)).

\noindent\textbf{Number of inpainting viewpoints.}
We finally analysis on the number of inpainting viewpoints in Fig. \ref{fig:ablation_imitate_inpaint}.
Different from previous works that use repetitive inpainting on perspective views to cover all the visible surfaces of the mesh, our method follows the principle that generates an informative panoramic texture and then propagates it through inpainting and imitating techniques.
Therefore, we don't rely on too many inpainting views, since inpainting itself cannot always produce reasonable images especially when observing occluded areas from small grazing angles (\eg, small gaps between the sofa and the floor).
As shown in Fig. \ref{fig:ablation_imitate_inpaint}, we don't observe significant improvement when increasing the number of inpainting views, as the first panoramic texture already endues sufficient appearance and overall impression of the indoor scenes.

\begin{figure}[!t]
\centering
\includegraphics[width=1.0\linewidth, trim={0 0 0 0}, clip]{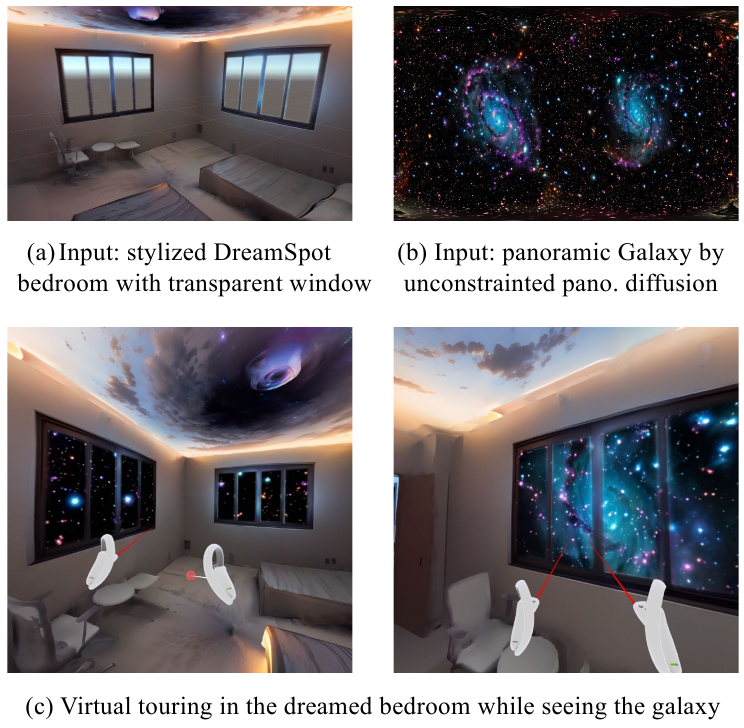}
\caption{
We build up a VR application by uploading textured scene assets with transparent windows and generated skyboxes into the HMD devices, which delivers an enchanting and immersive VR experience by allowing 6-DoF free-viewpoint touring with teleportation (red dot on the ground) in the fully stylized spaces.
    }
\label{fig:vr_app}
\end{figure}

\subsection{Immersive VR Application}
\label{ssec:exp_vr}

Once the stylized texture has been generated for the given scene mesh, we can directly place it into game engines such as Unity and upload it to the HMD devices for virtual touring.
To further improve the immersive experience, as shown in Fig. \ref{fig:vr_app}, we also make transparent windows on the user-defined region by assigning transparent alpha values on the baked UV images, where the UV space alpha mask is generated in a way similar to inpainting masks (Sec. \ref{ssec:method_texture_prop}).
Then, we pack the scene with an additional generated panoramic skybox by an unconstrained version of the panoramic diffusion model (\ie, the LDM in Sec. \ref{ssec:method_texture_gen} that trained on broaden equirectangular projection images).
During the rendering, we use the generated panoramic skybox as the background and open the virtual window with transparent UV textures.
In this way, we can build up a fantasy VR application, which allows users to enjoy the stylized space with their familiar scene structure but totally different appearance, \ie, seeing the nebula from the virtual window on a galaxy-theme bedroom.
Please refer to the supplementary video for the video recording of the immersive VR application.

\section{Conclusion}

We have proposed a novel text-driven indoor scene texturing framework, which enables to generate high-resolution and semantic meaningful UV textures for real-world scenes based on text prompts.
The key insight of our work is to first synthesize a stylized panoramic view of the scene that already conveys a global consistent appearance, and then propagate it to the rest regions.
For texture propagation, we design novel confidential inpainting and implicit imitating techniques, which properly handle cluttered real-world geometry and maintain spatial coherence for occluded areas or thin structures.
The resulting stylized textured mesh can be feasibly uploaded into HMD devices, which delivers immersive VR experiences.

\noindent\textbf{Limitations and future works.}
Despite the novel scene-texturing capability provided by our method, it still has some limitations.
First, the panoramic texture synthesized by our method already bakes the scene lighting effects, which cannot support custom lighting or dynamic shadows in the rendering pipeline.
Second, to ensure high-quality texturing and a completely immersive VR experience, our method requires the input reconstruction to include real-world textures, and also relies on the quality of the scene reconstruction (\eg, incomplete scanned scenes without a roof such as ScanNet~\cite{dai2017scannet} is not preferred).
Third, our method does not support extra large rooms (\eg, theater, church) or outdoor spaces, as such scenarios might need multiple partitioned stylized panoramas to fill the entire scene.
In the future, we plan to support PBR texturing by fine-tuning LDM with PBR-based equirectangular projections, which would be more compatible with modern physically based rendering pipelines.
Besides, we can also incorporate our scene texturing pipeline with a visual positioning system, so as to align the stylized scene with the physical real world on HMD devices, which could deliver appealing MR experiences.

\noindent\textbf{Acknowledgements.}
We thank Freepik for icons in the figures.

{\small
\bibliographystyle{ieee_fullname}
\bibliography{main}
}

\end{document}